# Irregular-Time Bayesian Networks

**Michael Ramati** and **Yuval Shahar**
Information System Engineering
Ben-Gurion University
Be'er Sheva, Israel
{ramatim, yshahar}@bgu.ac.il

## Abstract

In many fields observations are performed irregularly along time, due to either measurement limitations or lack of a constant immanent rate. While discrete-time Markov models (as Dynamic Bayesian Networks) introduce either inefficient computation or an information loss to reasoning about such processes, continuous-time Markov models assume either a discrete state space (as Continuous-Time Bayesian Networks), or a flat continuous state space (as stochastic differential equations). To address these problems, we present a new modeling class called *Irregular-Time Bayesian Networks* (ITBNs), generalizing Dynamic Bayesian Networks, allowing substantially more compact representations, and increasing the expressivity of the temporal dynamics. In addition, a globally optimal solution is guaranteed when learning temporal systems, provided that they are fully observed at the same irregularly spaced time-points, and a semiparametric subclass of ITBNs is introduced to allow further adaptation to the irregular nature of the available data.

## 1 INTRODUCTION

In many fields observations are performed irregularly along time, due to either a limitation of the measurement process or lack of a constant immanent rate. Thus, we are interested in a general efficient method to model, learn and reason about structured stochastic processes that produce qualitative and quantitative data at irregularly spaced time-points. Markov models that may be used to support these tasks differ in the way they represent time.

Discrete-time Markov models are not well designed for irregular time settings: Kalman filters, Hidden Markov Models, and Dynamic Bayesian Networks (DBNs) in general [Dean and Kanazawa, 1989, Murphy, 2002] require the specification of a constant time distance between each two consecutive observations. This requirement leads to computationally inefficient learning and inference when the modeled time granularity is finer than the time spent between consecutive observations, and to an information loss in the opposite case. In both cases, inference is limited to multiples of modeled time granularity and would otherwise require the learning of a new model.

Markov models that represent time continuously, on the other hand, handle well time irregularity but suffer from other limitations. Specifically, Continuous-Time Bayesian Networks [Nodelman et al., 2002] model state transitions and thus assume a discrete state space, whereas stochastic differential equations [Oksendal, 2003] assume a flat continuous state space, having no conditional independence among their solving subprocesses and typically sharing the same observation times among these subprocesses.

We present a new modeling class designed to better support probabilistic reasoning at irregular time settings. Specifically, *Irregular-Time Bayesian Networks* (ITBNs) generalize Dynamic Bayesian Networks such that each time slice may span over a time interval to accommodate interprocess delays, and time differences between consecutive slices may vary according to the available data and inference needs. This generalization precludes the different limitations that appear in continuous-time models (by not constraining the value space) and discrete-time models. Comparing to the latter, it allows a substantial reduction in the model size which leads to a corresponding increase in computational efficiency, and introduces greater expressivity of the temporal dynamics by a property that may be referred to as a dynamic- and stochastic- order Markovity. In addition, a globally optimal solution is guaranteed when learning temporal systems, provided they are fully observed at the same irregularly spaced time-points, and a semiparametric subclass of ITBNs is introduced to allow further adaptation to the irregular nature of the available data.

## 1.1 VARYING COEFFICIENT MODELS

Varying coefficient models are regression models that are linear in the regressors, but their coefficients are allowed to change smoothly with the value of other variables called effect modifiers [Hastie and Tibshirani, 1993]. Formally put using the conditional independence notation,

**Definition 1.** *Let $Y$ be a random variable and $\{X_i, U_i : 1 \leq i \leq p\}$ be its predictors. If there exist (unknown) functions $\{\beta_i : 1 \leq i \leq p\}$ such that*

$$Y \perp \{X_i, U_i : 1 \leq i \leq p\} \mid \sum_{i=1}^{p} \beta_i(U_i) X_i$$

*then $Y$ is said to have a Varying Coefficient Model and $U_i$ is called the effect modifier of $X_i$ for each $1 \leq i \leq p$.*

Varying coefficient models typically take the settings of generalized linear models [McCullagh and Nelder, 1989]. Estimation methods developed for varying coefficient models are mainly based on local regression, and the kernel method as a special case [Fan and Zhang, 2008]. Penalized splines [Ruppert et al., 2003] generalize smoothing splines to allow more flexible choices of the spline model, the basis functions for that model, and the penalty. Taking the approach of penalized splines, we model (omitting the index $i$):

$$\beta(U) = \sum_{j=0}^{d} \beta_j U^j + \sum_{k=1}^{\kappa} \beta_{k+d} b_{k,d}(U)$$

where $b_{k,d}(\cdot)$ stand for some basis function such as B-spline, or the truncated power function captured by:

$$b_{k,d}(U) = (U - u_k)_+^d = \begin{cases} 0 & , U \leq u_k \\ (U - u_k)^d & , U > u_k \end{cases}$$

In both cases, $\{u_k : 1 \leq k \leq \kappa\}$ are called the *knots* of the spline $\beta(\cdot)$ and require a choosing scheme, which by default can be the $\frac{k}{\kappa+1}$-th quantiles of the available data for $U$, where the number of knots $\kappa$ parametrizes the complexity of the model along with the degree $d$.

## 2 REPRESENTATION

Time is typically represented by the natural numbers $\mathbb{N}$ (in discrete-time models) or the real numbers $\mathbb{R}$ (in continuous-time models). In irregular time settings we are interested in only some of these time-points, so a choice to represent time as a subset of a typical time set (T $\subseteq \mathbb{N}$ or T $\subseteq \mathbb{R}$) may seem reasonable. However, this time subset is not always fixed ahead of sampling, nor shared among all sample paths.

We propose a new method to represent time as a random vector $(T_j : j \in \mathbb{N})$, indexed by the order of the time-points of interest. The fundamental notion of stochastic processes may then be generalized as follows to accommodate this representation method.

**Definition 2.** *Let T be a set. Any function $\vec{X}$ from T, such that $\vec{X}(t)$ is a random variable for each $t \in$ T, is called a stochastic process. If T is a random vector $(T_j : j \in \mathbb{N})$, then $\vec{X}$ is said to have an irregular time.*

The above is not to suggest that specific probabilistic models should be sought for time, but rather that only the time-points of interest need to be considered, and these should appropriately be treated as evidence (i.e., data). Clearly, irregular-time stochastic processes generalize discrete-time stochastic processes given the vector T. The following proposition shows that in irregular time settings this generalization may imply model size reduction.

**Proposition 3.** *Let $\mathbf{T} = (T_j : 1 \leq j \leq n)$ be the vector of random time-points at which a discrete-time process is sampled, and $\mathcal{U} = \{T_j - T_{j-1} : 2 \leq j \leq n\}$. If the members of $\mathcal{U}$ are independent, then the probability that a discrete time model corresponding $\mathbf{T}$ has at least $\max(\mathbf{T}) - \min(\mathbf{T})$ points is monotonic in $|\mathbf{T}|$, and strictly monotonic if additionally $\Pr(U = 1) > 0$ for each $U \in \mathcal{U}$. If the members of $\mathcal{U}$ are further identically geometrically distributed with success parameter $p$, then an irregular time model corresponding $\mathbf{T}$ is expected to have $1/p$ times less points as $n \to \infty$.*

*Proof.* A discrete time model has $\max(\mathbf{T}) - \min(\mathbf{T})$ points if its time granularity, given by $\gcd(\mathcal{U})$, is 1. If $\mathcal{U} = \mathcal{U}' \cup \{U\}$, then

$$\begin{aligned}
\Pr(\gcd(\mathcal{U}) = 1) &\geq \Pr(\gcd(\mathcal{U}') = 1 \vee U = 1) \\
&= 1 - \Pr(\gcd(\mathcal{U}') > 1 \wedge U > 1) \\
&= 1 - \Pr(\gcd(\mathcal{U}') > 1) \Pr(U > 1) \\
&\geq 1 - \Pr(\gcd(\mathcal{U}') > 1) \\
&= \Pr(\gcd(\mathcal{U}') = 1)
\end{aligned}$$

and $\Pr(\gcd(\mathcal{U}) = 1) = \Pr(\gcd(\mathcal{U}') = 1)$ only if $\Pr(\gcd(\mathcal{U}') = 1) = 1$ or $\Pr(U = 1) = 1 - \Pr(U > 1) = 0$. Since discrete time models require $\frac{\max(\mathbf{T}) - \min(\mathbf{T})}{\gcd(\mathcal{U})}$ points to avoid information loss, this number is expected to approach $n/p$ as $n$ increases. Irregular time models require only $n$ points, leading a size ratio of $1/p$. □

The above suggests that irregular-time models are potentially more compact than discrete-time models, and that this potential increases with the number of observations. The compact irregular-time representation naturally supports computational savings in both memory and time consumed by the learning and inference tasks, which will be further discussed in the next two sections. However, the introduction of irregular-time processes caters for additional less trivial benefits.

First, given a memory consumption, an irregular-time model potentially support a longer hindsight for learning its parameters than a discrete-time model, because it may contain less points. This longer hindsight is essential for the convergence of parameter learning the more variables exist, due to either model factorization or data incompleteness. Secondly and similarly, the more factored the model is and incomplete the data are, the greater the difference between the effective model complexity of a discrete-time model and that of an irregular-time model. This in turn may lead to a better fit of the learned model.

Third, the ability to compute a probability given an evidence from the far past in one step implies, that long-distance effects can be directly expressed, as opposed to the case of regular time models. If we consider an irregular-time Markov model, this property could be referred to as both stochastic- and dynamic-order Markovity, because the distances between consecutive time-points depend on the data (and are thus not fixed) and may change over time (and are thus not static). Although this property may be interpreted as an irregular-time analog to higher-order Markovity in discrete-time models, these two properties are actually orthogonal. Like regular $p$-order Markov models, irregular $p$-order Markov models also consider the $p$ recent points for each present point, but unlike the former these points may be irregularly distanced.

Fourth, there is no need to choose a single temporal granularity and consequently no need to approximate the real time stamps, as may be necessary with discrete-time models. Lastly, irregular-time models support the application of filtering by demand, which may otherwise be computationally infeasible.

### 2.1 FACTORIZATION

To begin, we recall an underlying notion and notations.

**Definition 4.** Let $\mathcal{G} = (\mathcal{V}, \mathcal{E})$ be a directed acyclic graph such that $\mathcal{V}$ are random variables, and denote $\pi_\mathcal{G}(Y) = \{X : (X, Y) \in \mathcal{E}\}$ and $f_X(x) = \Pr(X = x)$. If $f_\mathcal{V}(\cdot) = \prod_{X \in \mathcal{V}} f_{X|\pi_\mathcal{G}(X)}(\cdot|\cdot)$ and $\Theta$ parametrizes $\{f_{X|\pi_\mathcal{G}(X)} : X \in \mathcal{V}\}$, then $(\mathcal{G}, \Theta)$ is called a *Bayesian Network* over $\mathcal{V}$.

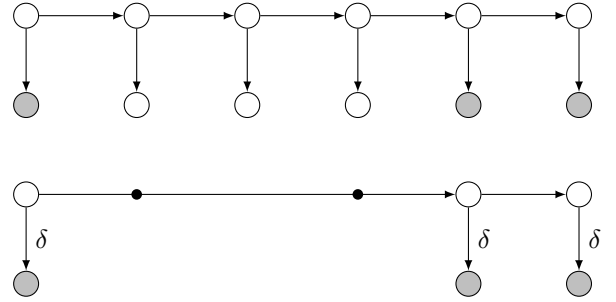

Figure 1: DBNs vs. ITBNs. Transparent circles represent unobserved states, shaded circles represent observed states, and opaque small circles represent knots. The DBN at the top has a constant time difference between consecutive slices. The ITBN at the bottom has time slices only at observation time-points, knots for changing dynamics between consecutive slices, and a delayed effect between nodes of the same time slice.

We now introduce a new factored model class, which orthogonally to random time differences, can express arbitrary delays in the effects of parent processes on child processes (using fixed time offsets $\delta_i$).

**Definition 5.** Let $\left(\vec{X}_i : 1 \leq i \leq m\right)$ be a vector of irregular-time processes, $(\delta_i \in \mathbb{R} : 1 \leq i \leq m)$ be a vector of time offsets, $\mathcal{V}_j$ denote $\left\{\vec{X}_i(T_j + \delta_i) : 1 \leq i \leq m\right\}$ for each $j \in \mathbb{N}$, and $(\mathcal{G}, \Theta)$ be a Bayesian network over

$$\bigcup_{j \in \mathbb{N}} (\mathcal{V}_j \cup \{T_{j+1} - T_j\})$$

If the following hold for each $1 \leq i \leq m$: $\pi_\mathcal{G}\left(\vec{X}_i(T_0 + \delta_i)\right) \subseteq \mathcal{V}_0$, $\pi_\mathcal{G}\left(\vec{X}_i(T_j + \delta_i)\right) \subseteq \mathcal{V}_{j-1} \cup \mathcal{V}_j$ for each $j \in \mathbb{N} \setminus \{0\}$, and $\Pr\left(\vec{X}_i(T_j + \delta_i) \mid \pi_\mathcal{G}\left(\vec{X}_i(T_j + \delta_i)\right), \Theta\right)$ is constant in $j$, then $(\mathcal{G}, \Theta)$ is called an *Irregular-Time Bayesian Network* (ITBN), $(\mathcal{V}_j : j \in \mathbb{N})$ is called its *(irregular) time slices*, and $(T_{j+1} - T_j : j \in \mathbb{N})$ is called its *(irregular) time differences*.

As implied by the definition above, (irregular) time slices of ITBNs are needed only at the points of interest. This relaxation comparing DBNs may be significant, since reasoning at even a single time slice is potentially hard. However, if we choose all $\delta_i = 0$ and $T_j = j$, then the ITBN for which this choice is made is a DBN. The fact that ITBNs generalize DBNs may also be implied by the facts that ITBNs are joint irregular-time processes much like DBNs are joint discrete-time processes, and irregular-time processes generalize discrete-time processes given T.

For concreteness, we present the following class of conditional probability distributions for ITBNs, that takes semiparametric settings of varying coefficient models. This class may enable an optimal exploitation and a natural adaptation of ITBNs to the available data. Specifically, a greater model complexity, manifested for example as the number of spline knots, may be attributed at the times where more data are available.

**Definition 6.** Let $\left\{\vec{X}_i : 1 \leq i \leq p\right\}$ and $\vec{Y}$ be stochastic processes. If there exist functions $\alpha, \beta$ and parameters $\{\gamma_i \in \mathbb{R} : 1 \leq i \leq p\}$ and $\{\delta_i \in \mathbb{R} : 1 \leq i \leq p\}$ such that $\vec{Y}(T_j)$ has a generalized linear model for each $j \in \mathbb{N} \setminus \{0\}$ with the predictor

$$\eta_j = \alpha(T_j - T_{j-1}) + \beta(T_j - T_{j-1})\vec{Y}(T_{j-1}) + \sum_{i=1}^{p} \gamma_i \vec{X}_i (T_j - \delta_i)$$

then $\vec{Y}$ is said to have an *irregular-time generalized linear* conditional probability distribution (CPD).

## 3 LEARNING

We now make the following important distinction.

**Definition 7.** Let $(\mathcal{V}_j : j \in \mathbb{N})$ be the (irregular) time slices of some ITBN. If there exists $n \in \mathbb{N}$ such that $\mathcal{V}_j$ is fully given for each $j \leq n$ and no $X \in \mathcal{V}_j$ is given for each $j > n$, then the data are said to be *irregularly complete* and the ITBN is said to be *fully observed*.

Clearly, complete temporal data in the regular notion implies irregularly complete data, and a similar implication applies to the notion of fully observed. However, no implication holds in the opposite direction. Specifically, a process that is fully observed at irregularly spaced time-points may be considered highly incomplete in the regular notion. In the rest of this section we assume the observation time vector **T** is given.

### 3.1 STRUCTURE

Learning the structure of an ITBN whose fixed time offsets $(\delta_i \in \mathbb{R} : 1 \leq i \leq m)$ between the nodes $\left(\vec{X}_i(T + \delta_i) : 1 \leq i \leq m\right)$ of a single (irregular) time slice are known or assumed is not different than learning the structure of any Bayesian network. The same applies to learning the structure of fully observed ITBNs, because in that case the fixed time offsets inside each (irregular) time slice can be deduced.

However, learning these fixed time offsets from irregularly incomplete data pose a challenge, because these offsets are arbitrary and continuous, such that learning cannot be reduced to finding the right connectivity between already existing nodes. Thus, we need to consider an hypothesis space and a scoring function for the unknown time offsets. A natural choice for the hypothesis space is the one spanned by the time offsets that already appear in the data. Similarly, a natural choice for the scoring function would be one that prefers the more compact candidate structures; that is, the least amount of new invented nodes.

An additional structural entity that does not appear in regular-time models is the set of knots. The general heuristics mentioned in Section 1.1 support a default selection of knots for each marginal process given the number of knots for that process. This number manifests the complexity of the process, and further methods exist for selecting it in penalized splines. The adaptation of these algorithms, which optimize standard fit criteria over hypothesized numbers of knots, to spline-based ITBNs is thus straightforward.

### 3.2 PARAMETERS

Learning the parameters of ITBNs does not require a specific choice of a time granularity or constant observation rate. Additionally, the representation compactness supports a longer hindsight for learning, and potentially lead to a lower probabilistic model complexity, which in turn may introduce a better fit of the learned model. Specifically, a semiparametric approach supports an optimal exploitation and a natural adaptation of the model to the available data. An illustration of typical settings for learning ITBNs with semiparametric CPDs appears in Figure 2.

Another qualitative merit can be stated and proved as follows. Recall, the notion of fully observed models has a special interpretation in the context of ITBNs.

**Proposition 8.** *Parameter learning of a fully observed ITBN has a globally optimal solution.*

*Proof.* Let there be $s$ independent sample paths, and for each $1 \leq i \leq s$ let $\mathcal{B}_i = (\mathcal{G}_i, \Theta)$ be the ITBN for the $i$-th sample path and $(\mathcal{V}_{ij} : j \in \mathbb{N})$ be its (irregular) time slices, such that $X_{ijk} = x_{ijk}$ for each $X_{ijk} \in \mathcal{V}_{ij}, 1 \leq j \leq n_i, 1 \leq i \leq s$. The likelihood of the (shared) parameters of $\{\mathcal{B}_i : 1 \leq i \leq s\}$ is given by

$$\mathcal{L}(\Theta) = \Pr\left(\bigwedge_{i=1}^{s}\bigwedge_{j=1}^{n_i}\bigwedge_{k=1}^{m} X_{ijk} = x_{ijk} \mid \bigwedge_{i=1}^{s}\mathcal{B}_i\right)$$

$$= \prod_{i=1}^{s}\prod_{j=1}^{n_i}\prod_{k=1}^{m} \Pr(X_{ijk} = x_{ijk} \mid \pi_{\mathcal{G}_i}(X_{ijk}), \Theta)$$

Thus, to maximize $\mathcal{L}(\Theta)$ we need only to regress $\{X_{ijk} : 1 \leq i \leq s, 1 \leq j \leq n_i\}$ on $\{\pi_{\mathcal{G}_i}(X_{ijk}) : 1 \leq i \leq s, 1 \leq j \leq n_i\}$ for each $1 \leq k \leq m$. □

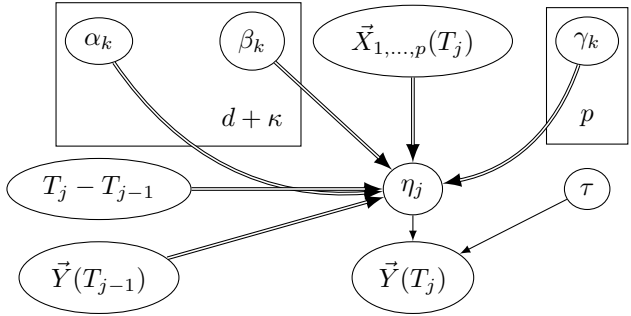

Figure 2: A typical irregular-time Gaussian linear CPD in Bayesian settings. Circles represent variables, rectangles represent replications, single-line arrows represent random relations, and double-line arrows represent fixed relations. $\vec{Y}(T_j)$ has a mean $\eta_j$ and a precision $\tau$. The vectors $\alpha, \beta$ parametrize the varying coefficients of $\eta_j$, each constitutes a $d$-degree and $\kappa$-knot penalized spline. $\tau$ is the constant precision parameter. The vector $\gamma$ constitutes the effects of the inside time-slice parents $\vec{X}_1(T_j), \ldots, \vec{X}_p(T_j)$ on $\eta_j$; we assume here for simplicity that $\delta = 0$.

## 4 INFERENCE

In this section we remove the assumption of fully given observation times taken at the last section. Thus, inference on ITBNs may refer to either of two basic tasks: estimating the unobserved states, or finding the time of an (at least partially) observed slice.

### 4.1 ESTIMATING THE STATES

Since ITBNs extend Bayesian networks, standard inference methods apply to ITBNs as well. However, since the definition of ITBNs implies there is no requirement for a constant time granularity, (irregular) time slices need to be constructed only at the time-points of specific interest. Specifically, if $\vec{X}$ has an irregular-time generalized linear CPD, then its computation given its parents, one of which at a randomly distanced time slice, is bounded by the number of knots between these slices.

Since ITBNs may have less time slices than DBNs, the same inference algorithms potentially yield a better performance. For example, denote $h$ the number of hidden nodes in a time slice; the (exact) transition from one time slice to the next may take $O(k^{2h})$ in discrete state space DBNs (where $k$ is the maximum number of states a hidden node can take), or $O(h^2)$ in general linear Gaussian DBNs [Murphy, 2002]. Less time slices clearly means less transitions, which may imply a better performance.

Specifically, computational savings are expected to take place when applying the task of smoothing, where states are estimated given their past and future, because inference may then involve many time slices. Prediction, where future states are estimated given the present, is beneficial with ITBNs especially when the time-point of interest is far (in terms of a regular observation rate) in the future. The filtering task, where the current state is typically estimated online, may be applied to ITBNs in a lazy evaluation fashion when only new data arrive or a special interest is paid. A similar application with discrete-time models may be infeasible.

### 4.2 FINDING THE TIME

Given two consecutive time slices, we may be interested to know in what point of the time interval they span, a third (partial or full) slice is likely to take place. Since we attribute no probability model to the time difference nodes, one way to answer this question would be by applying a uniform discretization to the time between the two given time slices, and then estimation of the states at each time slice in that time interval. However, in case a satisfactory solution requires the discretization to be fine relatively to the length of the time interval, this solution may be quite costly and an alternative more efficient way should be sought.

We suggest to reduce this problem to stochastic root finding [Chen and Schmeiser, 2001] by subtracting the requested value in each node of the additional slice from the estimated value of that node. Stochastic root finding algorithms are based on the fixed nonlinear root finding algorithms [Brent, 1975], and handle the randomness of the function in question using simulation. Note, that this function may similarly refer to the estimated process mean, variance, quantile, or derivative in case of continuous-valued processes, and state probability in case of discrete-valued processes.

The following proposition states the conditions and implies the method to reduce finding the time in irregular-time Bayesian networks to the problem of stochastic root finding.

**Proposition 9.** Let $(\mathcal{G}, \Theta)$ be an irregular-time Bayesian network over $\{X_i(T_j) : 1 \leq i \leq m, 1 \leq j \leq n\}$, and $j \in \mathbb{N}$ such that $T_{j-1} = t^-$, $T_{j+1} = t^+$ and $T_j$ is hidden. Denote $g_{ij}(x, \cdot, t) = f_{X_i(T_j)|\pi_\mathcal{G}(X_i(T_j)), T_j}(x|\cdot, t)$ and $x_{i,j} = E(X_i(T_j)|T_j)$. If $g_{ij}(x, \cdot, t)$ are continuous in $t \in (t^-, t^+)$ for each $i \in \{1, \ldots, n\}$ and $x \in \mathbb{R}$, then for each $x$ between $x_{i,j-1}$ and $x_{i,j+1}$ there exists $t \in (t^-, t^+)$ such that $x = x_{i,j}$.

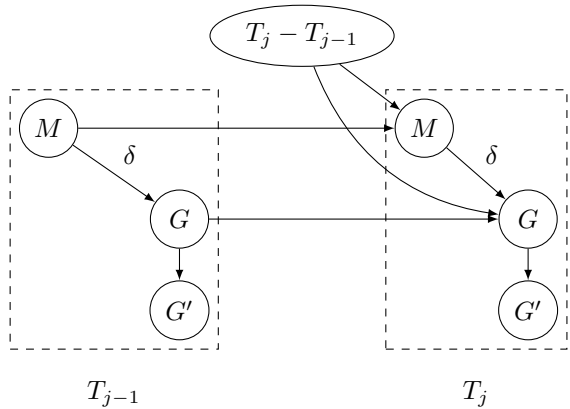

Figure 3: An ITBN. Glucose levels ($G$) are observed ($G'$) and depend on the time since the last meal has been started ($M$) minus the time to absorption in the blood ($\delta$). Both the measurements and the meals may occur at irregularly spaced time-points ($T$).

*Proof.* $\mathrm{E}\left(X_i\left(T_j\right)|T_j = t\right) = \int x f_{X_i(T_j)|T_j}(x|t)\, dx$ and $f_{X_i(T_j)|T_j}(x|t) = \int \cdots \int g_{ij}(x,\cdot,t) \prod_{k \neq i \vee l \neq j} g_{kl}(x_{kl},\cdot,t_l)\, dx_{kl}$ for each $x, t, t_l$ given $T_l = t_l, l \neq j$. Let $h_{ij}(t) = \mathrm{E}\left(X_i\left(T_j\right)|T_j = t\right) - x$ such that $x$ is between $x_{i,j-1}$ and $x_{i,j+1}$. Thus, $h_{ij}$ is continuous and has a root $t \in (t^-, t^+)$ due to the intermediate value theorem. □

## 5 APPLICATION

An R [Team, 2009] package has been created to learn and infer with generalized linear ITBNs using an interface to BUGS [Lunn et al., 2000, Thomas et al., 2006], and has been successfully tested on several irregularly sampled univariate processes. The purpose of this section is to highlight general qualitative properties of ITBNs, including their expressivity.

We demonstrate the fitting of the ITBN depicted at Figure 3 to the data on glucose levels over time [Hand and Crowder, 1996] plotted at Figure 4. The data contain 378 records of glucose levels collected for six subjects at irregularly spaced time points in reference to the starting time of their last meal. The ITBN learning scheme chooses the appropriate knot locations given the requested model complexity expressed in the number of knots per variable.

Figure 5 illustrates the estimated densities of the parameters that control the ITBN for the Glucose data and its discrete-time analog model. It is clear from this illustration that data with multi-modal parameters in discrete-time models may have unimodal fits with irregular-time models, which means greater likeli-hood. In addition, although the DBN has three times less CPD parameters than the ITBN has, it also requires 1640 unobserved variables comparing 376 in the ITBN, so its effective complexity is worse.

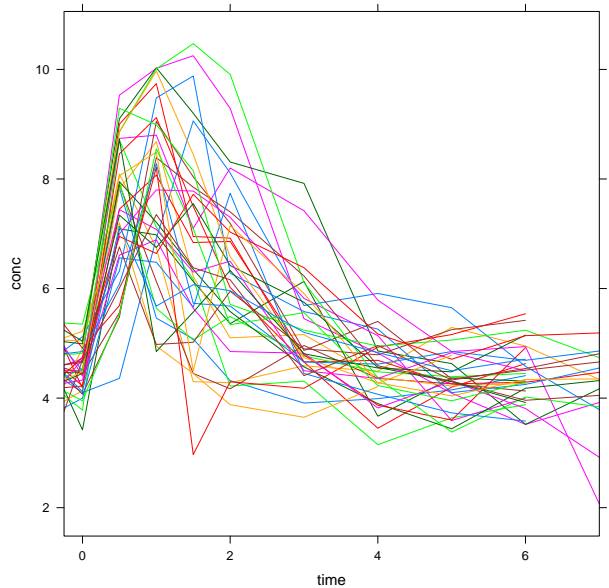

Figure 4: Glucose levels vs. time since the last meal. Each line connects successive observations per patient.

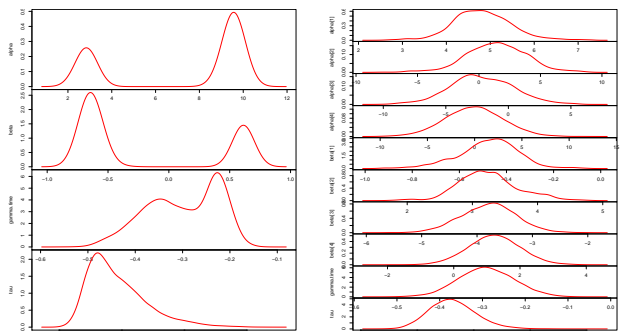

Figure 5: Density plots for the parameters of the Glucose DBN (on the left) and 2-knot ITBN (on the right).

## 6 RELATED WORK

Quantitative Temporal Bayesian Networks [Colbry et al., 2002] augment standard DBNs with Time Nets [Kanazawa, 1991] to allow for the representation of qualitative time intervals over the occurrence of events. The problem of learning these models is not addressed. Solutions to stochastic differential equations were suggested [Wilkin and Nicholson, 2000] to serve as CPDs in DBNs, but the issue of learning such CPDs was not addressed nor the integration with regular CPDs.

Continuous-Time Particle Filter [Ng et al., 2005] is an online simulation-based inference method that accounts for an hybrid state space, but continuous-valued subprocesses cannot factor others, as they solve (dependent) stochastic differential equations.

Recently, the integration of Bayesian and graphical modeling with semiparametric regression has been studied [Teh and Jordan, 2009, Wand, 2009, Fahrmeir and Raach, 2007], but the research interaction of these two disciplines seems to have been working mainly in one direction. So far, this integration efforts have yielded only extensions to statistical regression models, whereas belief models are not known to have enjoyed the semiparametric approach.

In longitudinal data analysis [Fitzmaurice, 2008], the semiparametric approach is used as one alternative to Markov models that are typically restricted to exponential correlation structures in irregular time settings. Specifically, Time-Varying Coefficient models [Hoover et al., 1998] use time to modify the parameters that control them, rather than their homogeneously parametrized states.

# 7 CONCLUSIONS

We have introduced Irregular-Time Bayesian Networks, a new modeling class that generalizes DBNs to better support inference and learning at irregular time settings. Specifically, each time slice may span over a time interval to accommodate fixed delays in the effects of parent processes on child processes, and time differences between consecutive slices may vary according to the available data and inference needs.

Several issues should be pointed here. Although the need to choose a specific time granularity has been lifted, the number of knots may still need to be sought. Learning irregular-time non-Gaussian CPDs may be intractable, and standard approximation methods should then be applied. In addition, however small and adaptive the number of time slices may be, leading to fewer inference transitions and guaranteeing a globally optimal learning under weaker conditions, each time slice still needs to be fully estimated. Finally, ITBNs allow long-distance effects to be directly expressed, but the representation method of time as a random vector may introduce several methodological questions, as follows.

First, whether time is really believed to "jump" from one observation to the next. Such an interpretation attributes ontological rather than epistemological semantics to the irregular-time representation. We do not argue for such true semantics, but only for the potential effectiveness of this modeling, following Box's pragmatic modeling philosophy ("all models are wrong, but some work").

Second, whether a sample path should affect the model of the underlying process that produced it; in other words, whether the sampling process should affect the modeling of the sampled process. Clearly, the sampling process has already a great affect on fitting the model of the sampled process, however implicit this affect may be. Thus, modeling this affect may be regarded as self, introspective, reflective Bayesian. However, we are interested in such modeling only to the extent it serves its inference purposes.

Third, how adding or removing fully unobserved time slices are prevented from affecting the predictions, potentially leading to inconsistent inference; in other words, how modifications to the time model maintain the same inferred results. This question masks the belief in a true time model which is not always a pragmatic belief. We note that changing the time granularity in discrete-time models clearly has similar effects.

Fourth, whether DBNs generalize ITBNs by having the time difference as either regular nodes or CPDs. ITBNs indeed generalize DBNs as has been shown, but the generalization goes in only one direction. Time differences cannot serve as regular nodes in DBNs, because they are not being attributed with any prior probability model and they belong to no time slice. Time differences also cannot serve as CPDs in DBNs, because they are random.

There are several directions into which this paper may be continued. One, adapting variational methods to ITBNs and conducting an appropriate evaluation. Another, studying the application of the ideas presented here to undirected graphical models, non-Markovian models, and Markov decision models. We also believe these ideas may be used to generalize object-oriented, relational, and first-order extensions to DBNs.

This paper has proposed a new method to represent time, potentially leading to improvements with respect to inference computational complexity, model probabilistic complexity, dynamics' expressivity, and adaptivity of the model to the data. To the best of our knowledge, this paper is also the first to suggest any application of a semiparametric approach to belief models, rather than an extension of statistical models with a graphical or Bayesian approach.

**Acknowledgments**

The authors would like to thank the anonymous referees for their constructive reviews, as well as the program chairs for facilitating these reviews.